%% file: iclr2020_conference.tex
\documentclass{article} 
\usepackage{iclr2020_conference,times}

\input{math_commands.tex}

\usepackage{hyperref}
\usepackage{url}
\usepackage{booktabs}       
\usepackage{colortbl}
\usepackage{multirow}
\usepackage{graphicx}
\usepackage{float}
\usepackage{hyperref}
\usepackage[below]{placeins}

\title{Semantic Enrichment of Nigerian Pidgin \\ English for Contextual Sentiment \\ Classification.}

\author{Wuraola Fisayo Oyewusi \\
Data Science Nigeria\\
Lagos, Nigeria \\
\texttt{wuraola@datasciencenigeria.ai} \\
\And
Olubayo Adekanmbi \ \\
Data Science Nigeria \\
Lagos Nigeria \\
\texttt{olubayo@datasciencenigeria.ai} \\
\AND
Olalekan Akinsande \\
Data Science Nigeria
Lagos,Nigeria \\
\texttt{olalekan@datasciencenigeria.ai}
}

\iclrfinalcopy 
\begin{document}

\maketitle

\begin{abstract}
Nigerian English adaptation, Pidgin, has evolved over the years through multi-language code switching, code mixing and linguistic adaptation. While Pidgin preserves many of the words in the normal English language corpus, both in spelling and pronunciation, the fundamental meaning of these words have changed significantly. For example, ‘ginger’ is not a plant but an expression of motivation and 'tank' is not a container but an expression of gratitude.
The implication is that the current approach of using direct English sentiment analysis of social media text from Nigeria is sub-optimal, as it will not be able to capture the semantic variation and contextual evolution in the contemporary meaning of these words. In practice, while many words in Nigerian Pidgin adaptation are the same as the standard English, the full English language based sentiment analysis models are not designed to capture the full intent of the Nigerian pidgin when used alone or code-mixed.
By augmenting scarce human labelled code-changed text with ample synthetic code-reformatted text and meaning, we achieve significant improvements in sentiment scoring.
Our research explores how to understand sentiment in an intrasentential code mixing and switching context where there has been significant word localization.This work presents a 300 VADER lexicon compatible Nigerian Pidgin sentiment tokens and their scores and a 14,000 gold standard Nigerian Pidgin tweets and their sentiments labels.

\end{abstract}

\section{Background}
Language is evolving with the flattening world order and the pervasiveness of the social media in fusing culture and bridging relationships at a click. One of the consequences of the conversational evolution is the intrasentential code switching, a language alternation in a single discourse between two languages, where the switching occurs within a sentence \citep{r1}. The increased instances of these often lead to changes in the lexical and grammatical context of the language, which are largely motivated by situational and stylistic factors \citep{r2}. In addition, the need to communicate effectively to different social classes have further orchestrated  this shift in language meaning over a long period of time to serve socio-linguistic functions \citep{r3}
Nigeria is estimated to have between three and five million people, who primarily use Pidgin in their day-to-day interactions. But it is said to be a second language to a much higher number of up to 75 million people in Nigeria alone, about half the population.\citep{tosin}. It has evolved in meaning compared to Standard English due to intertextuality, the shaping of a text's meaning by another text based on the interconnection and influence of the audience's interpretation of a text. One of the biggest social catalysts is the emerging urban youth subculture and the new growing semi-literate lower class in a chaotic medley of a converging megacity \citep{r5} \citep{r6}
VADER (Valence Aware Dictionary and sEntiment Reasoner) is a lexicon and rule-based sentiment analysis tool that is specifically attuned to sentiments expressed in social media and works well on texts from other domains. VADER lexicon has about 9000 tokens (built from existing well-established sentiment word-banks (LIWC, ANEW, and GI) incorporated with a full list of Western-style emoticons, sentiment-related acronyms and initialisms (e.g., LOL and WTF)commonly used slang with sentiment value (e.g., nah, meh and giggly) ) with their mean sentiment rating.\citep{hutto}.
Sentiment analysis in code-mixed text has been established in literature both at word and sub-word levels \citep{r7} \citep{r8} \citep{r9}. The possibility of improving sentiment detection via label transfer from monolingual to synthetic code-switched text has been well executed with significant improvements in sentiment labelling accuracy (1.5\%, 5.11\%, 7.20\%) for three different language pairs \citep{r6}

\section{Method}
This study uses the original and updated VADER (Valence Aware Dictionary and Sentiment Reasoner) to calculate the compound sentiment scores for about 14,000 Nigerian Pidgin tweets\footnote{Link to Nigerian Pidgin tweets and Sentiments \url{https://git.io/JvHrp}.}. The updated VADER lexicon (updated with 300 Pidgin tokens\footnote{Link to 300 Nigerian Pidgin Sentiments and Scores \url{https://git.io/Jv9og}.} and their sentiment scores) performed better than the original VADER lexicon. The labelled sentiments from the updated VADER were then compared with sentiment labels by expert Pidgin English speakers.

\begin{figure}[ht]
    \centering
    \includegraphics[width=13cm, height = 4cm]{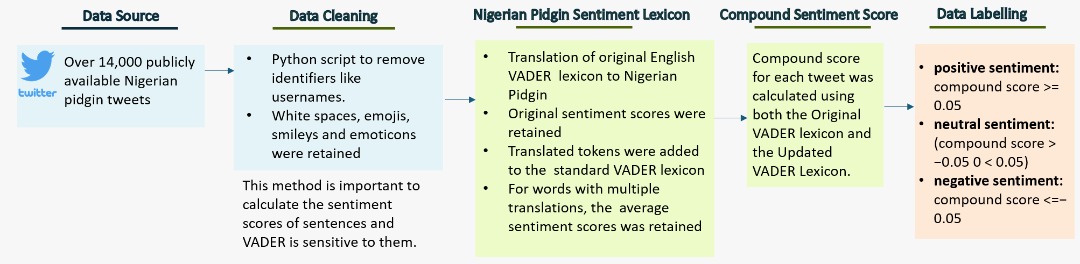}
    \caption{The semantic enrichment of Nigerian pidgin English for contextual sentiment classification methodology.
}
    \label{fig:galaxy}
\end{figure}
\FloatBarrier

\section{Results}
During the translation of VADER English lexicon to suitable one-word Nigerian Pidgin translation, a total of 300 Nigerian pidgin tokens
were successfully translated from the standard VADER English lexicon.
One of the challenges of this translation is that the direct translation of most the sentiment words in the original VADER English Lexicon translates to phrases not single one-word tokens and certain pidgin words translates to many english words.\ref{tab:scoreaverage}.

\begin{table}[H]

\caption{\label{tab:sentimentcompare}Nigerian Pidgin tweets with different sentiment labels}
\centering
\begin{tabular}[t]{>{\raggedright\arraybackslash}p{7em}>{\raggedleft\arraybackslash}p{7em}>{\raggedleft\arraybackslash}p{7em}>{\raggedright\arraybackslash}p{7em}>{\raggedright\arraybackslash}p{7em}>{\raggedright\arraybackslash}p{7em}}
\toprule
\textbf{Pidgin Sentence} & \textbf{Compound Sentiment Score before VADER English Lexicon Update} & \textbf{Compound Sentiment Score after VADER English Lexicon Update} & \textbf{Sentiment Label before VADER English Lexicon Update} & \textbf{Sentiment Label after VADER English Lexicon Update} & \textbf{Sentiment Label by Expert Pidgin Speaker}\\
\midrule
\rowcolor{gray!6}  som teams get black, som get purple but no one fine reach our jersey wey blue & -0.1154 & 0.7964 & negative & positive & positive\\
tiri kondoooooooooooooo! Sabi striker dzeko tear net wit pellegrini assist! & 0.0000 & 0.5080 & neutral & positive & positive\\
\rowcolor{gray!6}  gooooooooooooal!!! leonardo spinazzola throw beta cross enta and davide biraschi score for inside hin own post! 0-2 & 0.0000 & 0.6209 & neutral & positive & positive\\
\rowcolor{gray!6}  39’ willian try make beta pass, na beg we dey. & 0.0000 & 0.5106 & neutral & positive & positive\\
\addlinespace
Na to delete am & 0.0000 & -0.6908 & neutral & negative & negative\\
\rowcolor{gray!6}  Abed share your insight with me & 0.2960 & 0.5423 & positive & positive & positive\\
Why greenwood dey play nw?ole you don start & 0.3400 & -0.2500 & positive & negative & negative\\
\bottomrule
\end{tabular}
\end{table}

\section{Conclusion}
The quality of sentiment labels generated by our updated VADER lexicon is better compared to the labels generated by the original VADER English lexicon.\ref{tab:sentimentcompare}.Sentiment labels by human annotators was able to capture nuances that the rule based sentiment labelling could not capture.More work can be done to increase the number of instances in the dataset.

\bibliography{iclr2020_conference}
\bibliographystyle{iclr2020_conference}

\appendix
\section{Appendix}
\begin{table}[!h]

\caption{Average Sentiment Score for Nigerian Pidgin Sentiments with Multiple English Meanings}
\centering
\begin{tabular}[t]{llc}
\toprule
\textbf{Pidgin Words} & \textbf{VADER Sentiment Token and Score} & \textbf{Average Score}\\
\midrule
kasala & riot(-2.6), riots(- 2.), trouble(-1.7) & -2.2\\
gbege & catastrophe (3.4), chaos (2.7), chaotic(-2.2), problem (1.7), problems(-1.7) & -2.9\\
para & angry(-2.3), annoyed(-1.6), rage(-2.6) & -2.2\\
\bottomrule
\end{tabular}
\label{tab:scoreaverage}%
\end{table}

\subsection{Selection of Data Labellers}
Three people who are indigenes or lived in the South South part of Nigeria, where Nigerian Pidgin is a prevalent method of communication were briefed on the fundamentals of word sentiments.
Each labelled Data point was verified by at least one other person after initial labelling.

\subsubsection*{Acknowledgments}
We acknowledge Kessiena Rita David,Patrick Ehizokhale Oseghale and Peter Chimaobi Onuoha for using their mastery of Nigerian Pidgin to translate and label the datasets.

\end{document}

%% file: math_commands.tex
\usepackage{amsmath,amsfonts,bm}

\def\eqref#1{equation~\ref{#1}}

\def\1{\bm{1}}

\DeclareMathAlphabet{\mathsfit}{\encodingdefault}{\sfdefault}{m}{sl}
\SetMathAlphabet{\mathsfit}{bold}{\encodingdefault}{\sfdefault}{bx}{n}

